\documentclass[conference, 10pt]{IEEEtran}

\usepackage{cite}
\usepackage{amsmath,amssymb,amsfonts}
\usepackage{algorithm}
\usepackage{algpseudocode}
\usepackage{graphicx}
\usepackage{textcomp}
\usepackage{xcolor}
\usepackage{colortbl}
\usepackage{booktabs}
\usepackage{multirow}
\usepackage{tcolorbox}
\usepackage{hyperref}
\usepackage{subcaption}
\usepackage{enumitem}
\usepackage{balance}
\usepackage{listings}
\usepackage{xspace}
\usepackage{makecell}
\usepackage{pifont}
\usepackage{tikz}
\usetikzlibrary{positioning,shapes,arrows.meta}

\newcommand{\sys}{\textsc{SkillTrace}\xspace}
\newcommand{\bench}{\textsc{SkillTrace-Bench}\xspace}

\newcommand{\etal}{\textit{et al.}\xspace}

\begin{document}
\bstctlcite{IEEEexample:BSTcontrol}
\title{SkillTrace: Multi-Trace Provenance Auditing for LLM-Agent Skill Reuse}


\author{
\IEEEauthorblockN{
Jialuo Chen\IEEEauthorrefmark{1}\IEEEauthorrefmark{2},
Minghe Wang\IEEEauthorrefmark{2},
Lingqi Jiang\IEEEauthorrefmark{2},
Jianan Ma\IEEEauthorrefmark{3}\IEEEauthorrefmark{1},
Xinhao Deng\IEEEauthorrefmark{1}\IEEEauthorrefmark{4},\\
Xiaohu Du\IEEEauthorrefmark{1},
Ruixiao Lin\IEEEauthorrefmark{2},
Yunhao Feng\IEEEauthorrefmark{1},
Linkang Du\IEEEauthorrefmark{6},
Jingyi Wang\IEEEauthorrefmark{2}
}
\IEEEauthorblockA{
\IEEEauthorrefmark{1}Ant Group;
\IEEEauthorrefmark{2}Zhejiang University;
\IEEEauthorrefmark{3}Hangzhou Dianzi University, China;\\
\IEEEauthorrefmark{4}Tsinghua University;
\IEEEauthorrefmark{5}Xi'an Jiaotong University
}
}

\maketitle


\begin{abstract}
LLM-agent ecosystems are rapidly growing around reusable skills:
mixed-modality packages of metadata, natural-language instructions,
code, tools, references, and operational workflows. As skills become
marketplace artifacts, auditing their reuse is no longer the same
problem as ordinary code clone detection. Existing detectors target
single-modality source code or whole-package similarity, yet skill
reuse evidence is distributed across authored text, implementation
fragments, and operational structure. As a result, they can miss reuse that
preserves only one part of a skill.
We present \sys{}, a multi-trace provenance auditing framework for
LLM-agent skill reuse. \sys{} extracts three provenance traces: Expression,
Implementation, and Operational. It represents the Operational Trace
as a Skill Operational Graph (SOG) that captures activation,
procedure, and resource-flow structure. An LLM assists only
the Operational-trace extraction, once at ingestion; at audit time
\sys{} compares cached traces deterministically, calibrates each trace
against same-function strict negatives, and reports which trace supports
a reuse decision. On \bench{}, with 820 transformed reuse positives over
100 marketplace anchors and 751 negative controls, \sys{} achieves AUROC 0.938 and F1
0.898. A 36,446-skill wild audit further shows that
trace-attributed evidence surfaces actionable reuse review queues beyond
repository-level baselines.

\end{abstract}

\section{Introduction}\label{sec:intro}
LLM-agent ecosystems are rapidly growing around \emph{skills}:
reusable packages that let an agent acquire a capability by loading
natural-language instructions, executable snippets, tool interfaces,
reference files, and examples into its context~\cite{anthropic2026extend,
anthropic2026agentskillsdocs,anthropic2025skills}. Unlike a traditional
code artifact, a skill is mixed-modality and agent-facing: its behavior
is shaped not only by code, but also by instructions, activation cues,
tool-use rules, and workflow descriptions. As public registries and
marketplaces make skills easy to publish, adapt, and redistribute~\cite{ling2026agentskillsdatadrivenanalysis},
they are becoming software-like components whose reuse must be
understood and governed.

Reuse is not inherently harmful. Developers may fork a skill, adapt it to another host, extract a useful script, port a workflow across registries, or instantiate many skills from a common template.
The problem is provenance visibility: current catalogs mix independent implementations, template-generated variants, near-duplicates, and transformed derivatives that all appear as separate marketplace entries.
Recent ecosystem studies report substantial reuse and clone-like relations~\cite{kim2025toolcloning}, and show that exploitable flaws can propagate when vulnerable skills are copied, forked, or repackaged~\cite{agentskillswild2026}. 
Without provenance-aware reuse analysis, it becomes difficult to measure ecosystem growth, construct de-duplicated benchmarks, remediate propagated flaws, or maintain quality-aware registries. 
The underlying task is provenance auditing: given a reference skill and a candidate, decide whether the candidate inherits provenance-bearing artifacts from the reference rather than merely serving the same purpose.

Existing reuse and clone detectors fit this setting only partially. 
Source-code clone detectors such as MOSS~\cite{schleimer2003winnowing}, JPlag~\cite{prechelt2002jplag}, Deckard~\cite{jiang2007deckard}, and SourcererCC~\cite{sajnani2016sourcerercc} are designed for relatively uniform code substrates.
Repository-level fingerprinting methods such as sdhash~\cite{roussev2010sdhash} and ssdeep~\cite{kornblum2006identifying} compare whole-package byte or token content, and flat-text similarity collapses a skill into a single document representation. 
These techniques are useful for broad near-duplicate discovery, but skill reuse evidence may be distributed across authored text, implementation fragments, and operational design. 
A repository-level score cannot explain whether a candidate inherited a script, a workflow, or only a common scaffold.

\begin{figure*}[t]
\centering
\includegraphics[width=0.95\textwidth]{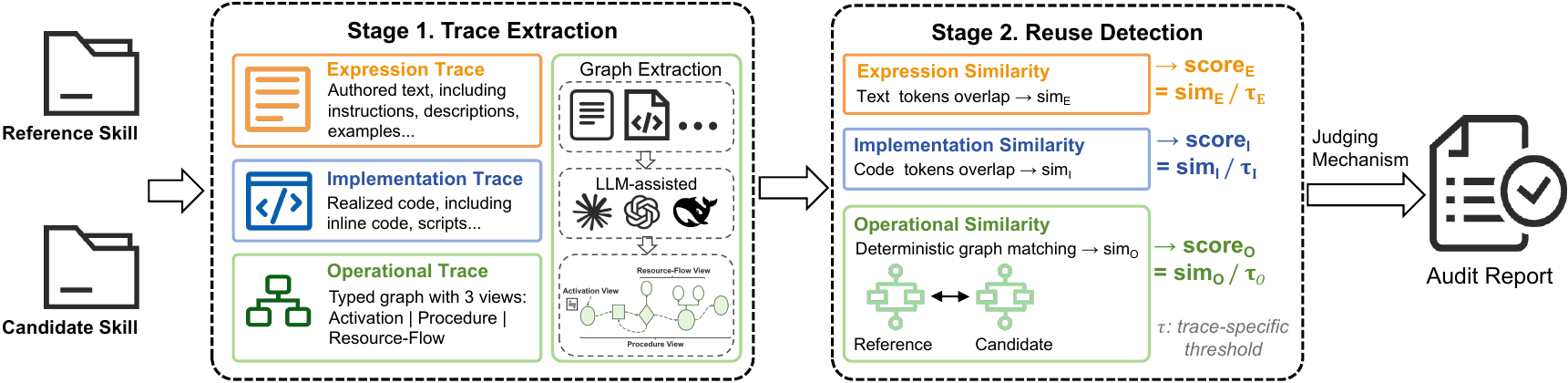}
\caption{\sys{} at a glance. The auditor extracts Expression,
Implementation, and Operational traces from reference and candidate
skill packages. The Operational Trace is represented as a Skill
Operational Graph (SOG) with Activation, Procedure, and Resource-Flow
views. At audit time, cached traces are compared deterministically and
\sys{} emits an audit report.}
\label{fig:overview}
\end{figure*}


The central challenge is therefore not whether two skills resemble each other globally, but which provenance survives transformation, and this is hard because a skill's reusable value is unevenly distributed. 
The value often concentrates in a few artifacts, such as a debugged support script, or a carefully designed tool-orchestration workflow, rather than in the surrounding boilerplate. 
A derivative can preserve these high-value artifacts while reorganizing files, rewriting documentation, renaming identifiers, porting the package to another host, or padding it with unrelated content.
Under such transformations global resemblance may be too weak to flag the relation even though concrete artifacts are inherited. 
Conversely, two independently developed skills that solve the same task may resemble each other globally since they share templates, APIs, or generic workflows. 
Skill reuse auditing must therefore recover transformed inheritance while controlling false attribution on same-function independent skills.

Our key insight is that skill reuse leaves complementary traces.
Expression evidence may vanish when descriptions, examples, or trigger prose are paraphrased, and implementation evidence may vanish when scripts, commands, or API calls are rewritten; yet operational evidence often remains: a derivative that preserves the skill's functionality must still retain much of its activation logic, task procedure, and resource flow.
Rewriting all three at once is possible, but it approaches the cost of building a new skill from scratch and so erodes the very value that motivated the reuse. 
A clean-room implementation of the same task, by contrast, may share high-level task semantics yet should not consistently preserve the same expression choices, implementation fragments, and operational structure. 
This motivates trace-attributed auditing: rather than producing one global resemblance score, an auditor should ask which provenance trace, if any, survives in the candidate.

Based on this insight, we present \sys{}, a trace-attributed framework for auditing reuse provenance in LLM-agent skills. 
As shown in Fig.~\ref{fig:overview}, \sys{} has two stages. 
During trace extraction, it extracts three provenance traces from each skill package: \emph{Expression}, capturing authored natural-language content; \emph{Implementation}, capturing executable realization such as scripts, commands, and API-use patterns; and \emph{Operational}, capturing host-aware execution design. 
The Operational Trace is represented as a Skill Operational Graph (SOG) with Activation, Procedure, and Resource-Flow views.
During reuse detection, \sys{} compares cached traces deterministically, calibrates each trace against same-function strict negatives, and emits a trace-attributed evidence report for human review. 
Expression and Implementation are computed deterministically, while an LLM is used only once at ingestion to normalize each skill's mixed-modality content into its Operational Graph; audits then run over cached traces with no further LLM calls.

To evaluate \sys{}, we construct \bench{}, a trace-preservation
benchmark over real marketplace anchors and same-function independent
controls. The benchmark includes transformed reuse positives covering
repackaging, porting or relocation across hosts, and partial
replacement; same-function strict negatives mined from the real corpus;
and metadata-only clean-room controls that probe the expected boundary
of artifact-based auditing. We further audit a 36,446-skill public
corpus to test whether \sys{} can build actionable reuse review queues
beyond controlled benchmarks.

Our contributions include:
\begin{itemize}
    \item We formulate post-hoc skill reuse provenance auditing as a
    software engineering problem over heterogeneous LLM-agent skill
    packages, separating inherited provenance-bearing traces from
    same-function independent similarity.

    \item We present \sys{}, a trace-attributed audit framework that
    extracts Expression, Implementation, and Operational traces. Its
    Operational Trace is represented as a Skill Operational Graph,
    enabling reviewable evidence about which part of a skill survived a
    transformation.

    \item We construct \bench{}, a benchmark with 820 transformed
    reuse positives over 100 real marketplace anchors and 751
    negative controls. \sys{} reaches AUROC 0.938 and F1 0.898.

    \item We conduct a 36,446-skill marketplace audit showing that
    trace-attributed evidence surfaces actionable reuse review queues,
    including trace-specific candidates that repository-level similarity
    under-prioritizes.
\end{itemize}

\section{Background}\label{sec:background}

\subsection{Software Reuse Detection}


Software reuse detection has long studied how to recognize copied,
adapted, or independently reimplemented code~\cite{roy2007clone}. Classical clone detectors
compare normalized token streams (MOSS~\cite{schleimer2003winnowing},
JPlag~\cite{prechelt2002jplag}, CCFinder~\cite{kamiya2002ccfinder},
SourcererCC~\cite{sajnani2016sourcerercc}), while tree- and
dependence-graph methods compare structure to resist local edits
(Deckard~\cite{jiang2007deckard}, GPLAG~\cite{liu2006gplag}). Fuzzy
digests and resemblance measures compare larger artifacts
(sdhash~\cite{roussev2010sdhash}, Broder~\cite{broder1997resemblance}),
and neural representations provide semantic code or text similarity
signals~\cite{guo2021graphcodebert,wang2021codet5,reimers2019sentencebert}.
A related idea is the \emph{software birthmark}: an intrinsic property
that tends to survive renaming, reformatting, and local refactoring
because removing it would change the program's behavior~\cite{schuler2007dynamic}.
Birthmarks are useful because they focus on behavior-bearing structure,
not only surface text. Skills need an analogous but broader artifact
model. A skill may preserve authored prose, executable fragments, or
host-facing operational design even when the rest of the package is
rewritten. \sys{} follows this lineage by treating Operational Trace as
a skill-level operational birthmark, while keeping Expression and
Implementation traces for textual and executable reuse evidence.

\subsection{LLM-Agent Skills and Tool Ecosystems}
LLM-agent capabilities are increasingly packaged as reusable artifacts~\cite{wang2024survey_agents,zhou2026skillgenbench,jiang2026sok}.
OpenAI's GPTs package custom instructions and knowledge for distribution
through the GPT Store~\cite{openai2023gpts,openai2024gptstore};
Anthropic's Agent Skills define filesystem packages centered on a
\texttt{SKILL.md} manifest with optional scripts and references
progressively disclosed to the agent~\cite{anthropic2025skills,
anthropic2026agentskillsdocs}; and the Model Context Protocol (MCP)
standardizes tool and data connections across agent platforms
~\cite{anthropic2024mcp,modelcontextprotocol2025spec,hasan2025model,zhao2025mcpattack}. Recent studies of agent skills
and tool ecosystems document the same shift toward reusable,
tool-mediated agent execution~\cite{ling2026agentskills,xu2026agentskills,
agentskillswild2026,liu2026maliciousskills}.
This packaging is now reflected in public and paid skill marketplaces.
Directories such as ClawHub\footnote{\url{https://clawhub.ai/}} and
SkillHub\footnote{\url{https://skillhub.club/}} index reusable skills
with creator and listing metadata, while paid marketplaces such as
Agensi\footnote{\url{https://www.agensi.io/}} support buying and selling
skill packages. This does not make reuse
inherently wrongful, but it makes provenance visibility operationally
important for registry maintenance, creator-facing review, and
marketplace governance.

The closest measurement work is Kim \etal's study of tool cloning in
agentic-AI ecosystems~\cite{kim2025toolcloning}, which measures
repository-level Jaccard and ssdeep similarity across MCP and Skills
repositories and manually validates high-similarity buckets. We use this
view as a deployable baseline: \emph{RepoClone} denotes our
implementation of Kim \etal's repository-level Jaccard/ssdeep similarity
over normalized skill packages. Our audit question is different: whether
a candidate skill preserves provenance-bearing traces under transformed
reuse, even when whole-package similarity is low.

\subsection{Prompt and Agent Artifact Protection}
Prompt-level protection is related but narrower. Watermarking schemes
such as PromptCARE~\cite{yao2024promptcare} and
PromptCOS~\cite{yang2025promptcos} protect prompts before release, while
prompt leaking, stealing, and injection work shows that proprietary
instructions can be exposed or abused in practice
~\cite{hui2024pleak,yang2025prsa,greshake2023indirect,liu2023houyi,perez2022ignore}.
Agent artifacts extend this problem beyond flat prompt text: skills
bundle instructions with code, tools, references, and operational
procedures. Prior protection mechanisms therefore complement rather than
replace post-hoc reuse auditing. \sys{} assumes two published packages
are available and asks whether observable provenance traces survive
marketplace transformations, without modifying or watermarking either
artifact.

\section{\sys{} Design}\label{sec:methodology}

\subsection{Problem Definition}
\label{sec:problem-definition}


We study \emph{post-hoc skill reuse auditing}. Given a reference
skill $R$ and a candidate skill $C$, the goal is to decide whether $C$ is derived from $R$ (a reuse candidate) or is an independently developed skill. 
A skill differs from a flat prompt or a code repository: it may
contain natural-language instructions, executable code, reference
documents and embedded workflows. Reuse may therefore preserve only part of the reference
skill. A candidate may inherit the reference wording, executable
implementation, or operational design while rewriting the rest of the
package. Conversely, two independently developed skills may solve the
same task and share APIs or templates while preserving no
provenance-bearing details.

We call $(R,C)$ a \emph{reuse candidate} when $C$ preserves at
least one concrete provenance trace of $R$ beyond the similarity
expected from same-function independent skills. \sys{} is not a legal
infringement detector; it produces calibrated, trace-attributed
evidence for review.

\begin{algorithm}[t]
\caption{\sys{} audit procedure}
\label{alg:skilltrace}
\begin{algorithmic}[1]
\Require Reference skill $R$, candidate skill $C$
\Ensure Reuse score and evidence report

\State $(\tau_E,\tau_I,\tau_O) \gets \textsc{LoadCalibratedThresholds}()$
\State $\beta \gets \textsc{LoadDecisionBound}()$
\Comment{default $\beta=1.0$}
\State $T_r \gets \textsc{ExtractTraces}(R)$
\State $T_c \gets \textsc{ExtractTraces}(C)$
\Comment{$T = (T_E, T_I, T_O)$}

\For{$k \in \{E,I,O\}$}
    \State $s_k \gets \textsc{TraceSimilarity}(T_r^k, T_c^k)$
    \State $z_k \gets s_k / \tau_k$
\EndFor

\State $z \gets \max(z_E, z_I, z_O)$
\State $\mathcal{E} \gets \textsc{CollectEvidence}(T_r, T_c, \{k : z_k \geq \beta\})$

\If{$z \geq \beta$}
    \State \Return $\textsc{Surfaced}(z, \mathcal{E})$
\Else
    \State \Return $\textsc{NotSurfaced}(z, \mathcal{E})$
\EndIf
\end{algorithmic}
\end{algorithm}

\subsection{Framework Overview}

\sys{} treats skill reuse as \emph{trace preservation}. Instead of
collapsing a skill pair into one whole-package similarity score, it
extracts three provenance traces and asks which trace, if any, remains
observable in the candidate.

The pipeline has two stages, shown in Fig.~\ref{fig:overview}. First,
during \textbf{1) Trace Extraction}, \sys{} parses each skill package once
and extracts an \emph{Expression Trace} for authored textual choices,
an \emph{Implementation Trace} for executable realization, and an
\emph{Operational Trace} for host-aware execution design. The
Operational Trace is represented as a Skill Operational Graph (SOG).
The extracted traces are cached and versioned, so repeated audits over
the same skill do not require re-extraction.
Second, during \textbf{2) Reuse Detection}, \sys{} compares the cached
traces of $R$ and $C$. Each trace produces a raw similarity score
and has its own threshold calibrated on same-function strict negatives.
A pair is surfaced when the calibrated MaxFusion score meets the
decision bound.
The output is an evidence report that records the firing trace, raw and
calibrated scores, and matched evidence pointers. Algorithm~\ref{alg:skilltrace}
summarizes the audit procedure.

\subsection{Multi-Trace Extraction}\label{sec:assets}
\sys{} extracts three provenance traces from each skill package:
Expression, Implementation, and Operational Trace (Table~\ref{tab:traces}).
They correspond to three ways in which reuse can survive rewriting. A
candidate may preserve the authored wording, the executable realization,
or the operational design of the reference skill. Keeping the traces
separate lets \sys{} report which part of the reference remains
observable instead of relying on a single whole-package score.

\subsubsection{Expression Trace}


Expression Trace captures \emph{how a skill is written}. It is extracted
from natural-language substrates: metadata, descriptions, when-to-use
fields, section prose, instructions, reference files, and assets. The extractor tokenizes these fields deterministically. Common stopwords, markdown boilerplate, and generic skill-scaffold terms
are removed.

Expression evidence is useful when a derivative keeps
documentation, examples, or language. Its limitation is also clear: prose can often be
paraphrased without changing the underlying skill behavior.

\subsubsection{Implementation Trace}

Implementation Trace captures \emph{how a skill is realized in
executable form}. It is extracted from inline code, support scripts, shell
commands, API calls, configuration schemas, and command-line interfaces.
Following token-based code clone detection~\cite{kamiya2002ccfinder,sajnani2016sourcerercc}, \sys{} uses a language-agnostic
token representation as the default implementation trace. This choice
matches skill packages, which often mix Python, JavaScript, shell
snippets, and natural-language command recipes. 

Implementation evidence is useful when a derivative keeps code
templates, scripts, API call patterns, or command recipes while
rewriting the surrounding documentation. It weakens when the code
is removed or rewritten in another language. 

\begin{table}[t]
\centering
\small
\caption{Three provenance traces extracted by \sys{}. Expression and Implementation adapt traditional text/code reuse signals to heterogeneous skill packages, while Operational Trace is represented as a Skill Operational Graph (SOG), a skill-level operational birthmark.}
\resizebox{\columnwidth}{!}{%
\begin{tabular}{@{}lll@{}}
\toprule
Trace & Primary substrate & Captures \\
\midrule
Expression
& NL text
& Authored wording, examples, trigger prose \\

Implementation
& Code / commands
& Scripts, API tokens, command interfaces \\

Operational
& Skill Operational Graph
& Activation, procedure, tool/resource flow \\
\bottomrule
\end{tabular}}

\label{tab:traces}
\end{table}

\subsubsection{Operational Trace}
Operational Trace captures \emph{how a skill works when hosted by an
agent}. This trace is specific to skills: a skill is not only text or
code, but also a package that must be discovered, activated, assembled
into context, and executed through tools and resources. \sys{} represents
this trace as a typed \emph{Skill Operational Graph} (SOG).

\subsection{Skill Operational Graph}
\label{sec:Operational-Graph}
Traditional software reuse detection often relies on code birthmarks:
structural properties that tend to survive renaming, formatting changes,
and local refactoring. Agent skills need a higher-level analogue because
their behavior is distributed across instructions, code snippets, tool
descriptions, reference files, activation rules, and host-specific
loading behavior. A derivative may therefore preserve the way the skill
operates even when its prose, file layout, or code identifiers have
changed.

\sys{} represents this higher-level identity as a \emph{Skill
Operational Graph}. The SOG is a skill-level operational birthmark: it
abstracts a skill into the actions it performs, the conditions under
which it is activated, and the tools, resources, and artifacts that flow
through its execution. It is inspired by control-flow and
data/dependence graphs~\cite{allen1970control,ferrante1987program}, but it is defined over agent-skill behavior
rather than program statements, variables, or code basic blocks.

Given a skill package $S$, the SOG is a typed graph
\begin{equation}
G_O(S) = (B_S, A_S, P_S, R_S).
\label{eq:sog-definition}
\end{equation}

where $B_S$ is the set of normalized operational blocks, $A_S$ is the
activation signature that describes how the skill is entered by an
agent host, $P_S$ is the procedure structure relating blocks through
ordering, dependency, branch, and fallback relations, and $R_S$ is the
resource-flow structure relating blocks to tools, resources, and
artifacts they consume or produce. These four parts answer four audit
questions: what operations exist, how the skill is entered, how the
operations are organized, and what tools or artifacts flow through them.

\textbf{Operational Block.}
An operational block is a normalized skill-level action unit
extracted from instructions, code, tool specifications, or reference
files. It describes one coherent operation in the skill, such as
validating an input, retrieving an external resource, transforming
data, invoking a tool, or producing an output. A basic block is a
unit of program execution; an operational block is a unit of skill
operation.

Each operational block is represented by five normalized fields as:
\[
(\textsf{action}, \textsf{object\_role}, \textsf{tool\_role},
  \textsf{inputs}, \textsf{outputs}).
\]
These fields capture the minimal reusable semantics of a skill
operation: what action is performed, what role of object it acts on,
what kind of tool is used, and which artifacts flow into and out of
the operation. Concrete names such as file paths, function names,
library identifiers, and example values are normalized away when
possible, because they are easy to rewrite; the operational roles are
harder to remove without changing the skill's behavior.

The SOG schema uses closed vocabularies for actions, object roles, tool
roles, activation contexts, and relation types. We derive these
vocabularies from corpus-level inspection and freeze them before extraction, merging rare lexical
variants into coarser roles. This keeps the Operational Trace abstract
enough to survive paraphrase and host relocation, but concrete enough to
avoid collapsing all same-function skills into a single task intent.

\begin{figure}[t]
\centering
\includegraphics[width=0.95\columnwidth]{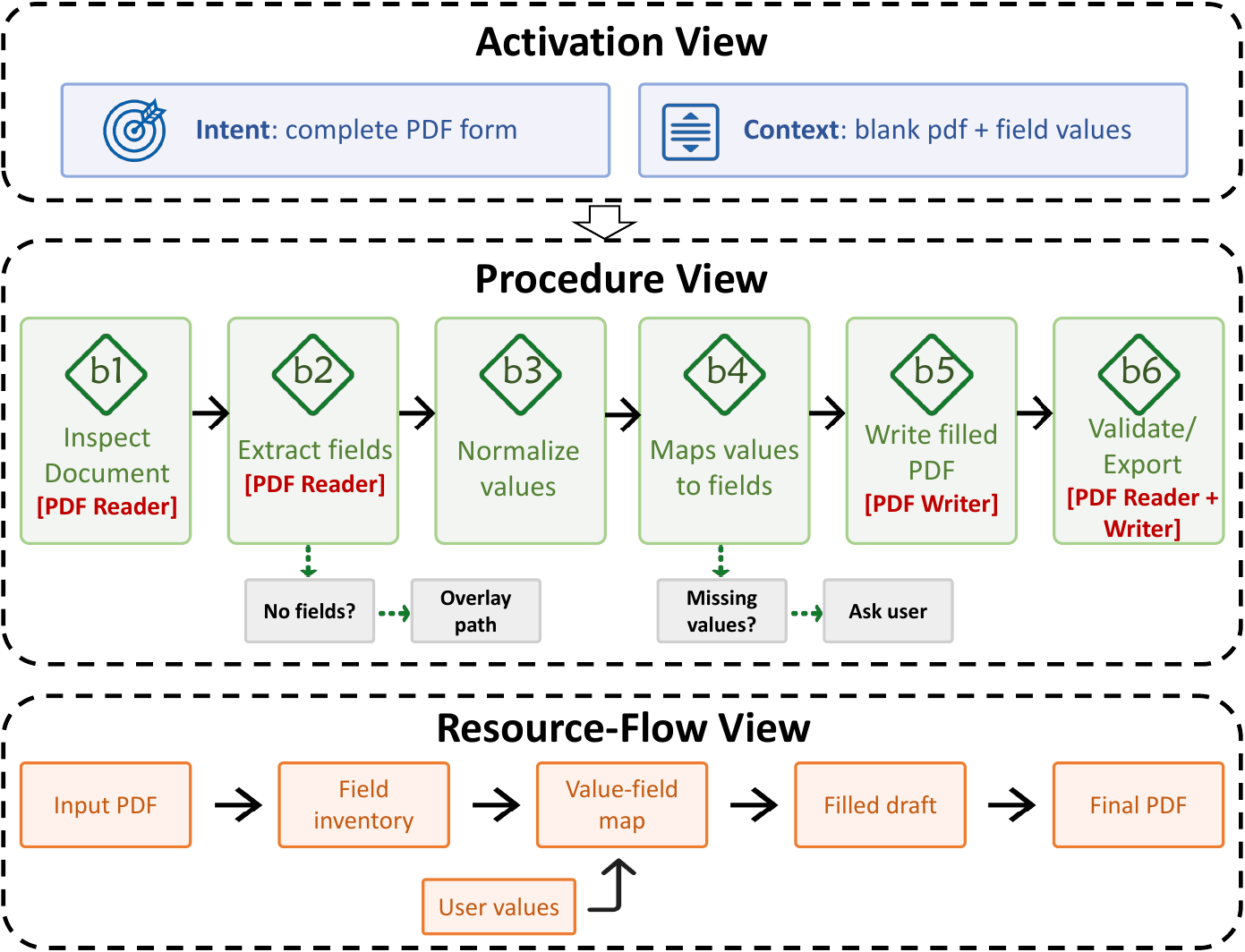}
\caption{Example Skill Operational Graph (SOG) for a PDF form-filling skill (simplified). }
\label{fig:exp_sog}
\end{figure}

\textbf{SOG views.}
The three SOG views expose different relations over the same operational
blocks. Activation describes how the graph is entered, Procedure
describes how blocks are ordered and conditioned, and Resource-Flow
describes how tools, resources, and artifacts connect to the blocks. 
Table~\ref{tab:sog_views} summarizes these views, and Fig.~\ref{fig:exp_sog} illustrates them using a PDF form-filling skill as an example.





\begin{table}[t]
\centering
\small
\caption{Internal views of the Skill Operational Graph.}
\resizebox{\columnwidth}{!}{%
\begin{tabular}{@{}lll@{}}
\toprule
View & Captures & Example relation \\
\midrule
Activation
& Entry context and start block
& \texttt{complete\_form} $\rightarrow$ \texttt{b1.inspect} \\

Procedure
& Order, branch, fallback
& \texttt{b1.inspect} $\rightarrow$ \texttt{b2.extract} \\

Resource-Flow
& Tool and artifact dependencies
& \texttt{pdf\_reader} + \texttt{blank\_form} $\rightarrow$ \texttt{b2.extract} \\
\bottomrule
\end{tabular}}
\label{tab:sog_views}
\end{table}


\emph{1) Activation View:} captures when and why the skill becomes
active: the user intent, required context, host-facing trigger
conditions, and the first operational blocks needed for that context.
It is analogous to a route-to-handler mapping but for agent skills.


\emph{2) Procedure View:} it captures how operational blocks are
sequenced: ordering, branches, fallback paths, failure handling, and
tool-call order. It is similar in spirit to a control-flow graph, but
its nodes are skill-level operations rather than code basic blocks.

\emph{3) Resource-Flow View:} it captures which tools, resources, and
artifacts flow through the procedure: support files, external
services, APIs, intermediate outputs, generated reports, and
validation artifacts. It is related to data-flow and dependence
graphs, but it operates at the level of agent resources and skill
artifacts rather than program variables.

These views are not separate detectors. They are projections of one
Operational Trace. Together, they explain how a skill can remain
operationally similar after a derivative rewrites wording, changes file
layout, or renames code identifiers.

\textbf{SOG Extraction}.
Operational traces are extracted once at ingestion time using an LLM
constrained to emit the SOG schema above. The extractor uses a
frozen schema and is instructed not to copy distinctive phrases from the
source package. The extracted graph is normalized, cached, and tagged
with the prompt and schema versions. Pairwise reuse detection does not
call an LLM: it compares cached SOG traces deterministically.

\subsection{Reuse Detection}

Once the traces have been extracted and cached, reuse detection becomes
a pairwise trace-comparison problem. 
Given a reference skill $R$ and a candidate skill $C$, \sys{} computes
one similarity score for each provenance trace:
\[
s_E(R,C), \quad s_I(R,C), \quad s_O(R,C).
\]
for Expression, Implementation, and Operational Trace respectively.
The scores are kept separate until calibration so that the final report
can identify which trace supports the reuse finding.


\subsubsection{Expression Similarity}

Expression similarity compares authored textual features. Let $T_E(S)$
be the normalized token set extracted from the natural-language
substrates of skill $S$. \sys{}
uses token Jaccard:
\begin{equation}
s_E(R,C)
=
\frac{|T_E(R) \cap T_E(C)|}
     {|T_E(R) \cup T_E(C)|}.
\end{equation}
Matched tokens are retained as evidence pointers for review.

\subsubsection{Implementation Similarity}

Implementation similarity compares executable and command-like
features. Following token-based code clone detection such as CCFinder and SourcererCC, \sys{} extracts normalized tokens
including inline code, support scripts and shell commands. Let $T_I(S)$ be
this implementation-token set. When the trace is applicable
on both sides, \sys{} computes:
\begin{equation}
s_I(R,C)
=
\frac{|T_I(R) \cap T_I(C)|}
     {|T_I(R) \cup T_I(C)|}.
\end{equation}
We use token-level similarity because skills often mix Python,
JavaScript, shell snippets, JSON schemas, API names, and natural-language
command recipes. A single AST representation would not cover this
heterogeneous setting. If either side lacks sufficient implementation
substrate, the Implementation Trace is marked as not applicable rather
than assigned a zero score.

\subsubsection{Operational Similarity}

Operational similarity compares the Skill Operational Graphs of two
skills. It asks four questions: do the skills contain similar
operational units; are those units entered through similar activation
conditions; are they organized by similar procedure structure; and do
they connect to similar tools, resources, and artifacts? As shown in \eqref{eq:sog-definition}, these four
questions correspond to the four parts of the SOG:

\[
\begin{aligned}
G_O(R)&=(B_R,A_R,P_R,R_R),\\
G_O(C)&=(B_C,A_C,P_C,R_C).
\end{aligned}
\]
\sys{} computes a deterministic similarity for each part and combines
them into one Operational score:
\begin{equation}
\begin{aligned}
s_O(R,C)=&
\lambda_B\,\mathrm{BlockSim}(B_R,B_C)\\
&+\lambda_A\,\mathrm{ActSim}(A_R,A_C)\\
&+\lambda_P\,\mathrm{ProcSim}(P_R,P_C)\\
&+\lambda_R\,\mathrm{ResSim}(R_R,R_C).
\end{aligned}
\label{eq:operational-similarity}
\end{equation}
We use fixed weights
\[
(\lambda_B,\lambda_A,\lambda_P,\lambda_R)
=(0.30,0.10,0.40,0.20),
\]
where the largest weight is assigned to procedure structure. These four terms are internal to Operational Trace; they are not
separate provenance traces.

\emph{BlockSim} captures whether two skills contain similar operational
units. Because a derivative may split, merge, rename, or relocate
operations, blocks are not compared position-by-position. Instead,
\sys{} uses maximum one-to-one matching between normalized operational
blocks, with a weak longest-common-subsequence term over the normalized
block sequence to capture coarse order preservation. \emph{ActSim}
compares compact activation signatures by normalized role overlap; it
is useful as host-facing corroboration, but is down-weighted because
same-function independent skills may share generic trigger conditions.
\emph{ProcSim} combines typed procedure-edge Jaccard with sequence
alignment over normalized block order, capturing dependencies, branches,
fallback paths, and execution order. \emph{ResSim} compares directed
typed block-resource dependency edges after node-label normalization;
resource flow is graph-structured but not globally ordered, because
tools, support files, intermediate artifacts, and outputs may branch,
merge, or be reused by multiple blocks.

This is a decomposed operational-graph match rather than full graph edit
distance. It keeps the graph information most relevant to skill reuse
while remaining deterministic, scalable, and easy to expose in an
evidence report.



\subsubsection{Calibrated Detection}
Raw trace scores are not directly comparable: Expression,
Implementation, and Operational Trace have different feature spaces and
different background collision rates. \sys{} therefore calibrates one
threshold per trace from same-function strict negatives:
\[
\tau_E, \quad \tau_I, \quad \tau_O.
\]
In our experiments, each $\tau$ is the 95th percentile of the
corresponding trace score on the strict-negative set, giving a
per-trace 5\% false-positive calibration without using positive
examples. In deployment, the same mechanism can be tightened or relaxed
to match a marketplace's review budget. Automatic threshold
selection is left as future work.


Let $\mathcal{A}(R,C)$ be the set of applicable traces for the pair
(e.g., Implementation is excluded when either side lacks enough
code/command substrate). For each applicable trace $i \in \{E,I,O\}$, \sys{} computes:
\begin{equation}
\mathrm{SkillTrace}(R,C)
=
\max_{i \in \mathcal{A}(R,C)}
\frac{s_i(R,C)}{\tau_i}.
\end{equation}
The pair is surfaced for review when this calibrated score meets or exceeds
a decision bound $\beta$. We use $\beta=1.0$ in all experiments, so one
trace meeting its N1-calibrated threshold is sufficient to enter the
review queue. In deployment, operators can raise $\beta$ for a smaller,
higher-confidence queue or lower it for exploratory triage. The maximum
rule matches the forensic setting: strong evidence in any one trace can
justify review even when other traces have been rewritten. A weighted
average would dilute such localized evidence and introduce tunable
cross-trace weights; calibrated max keeps the decision rule transparent.



\subsection{Interpretability}
\sys{} is designed to support human review. For every surfaced
pair, the report includes the firing trace, raw and calibrated scores,
and trace-specific evidence pointers: matched textual features for
Expression, matched code/API/command features for Implementation, and
matched SOG evidence for Operational Trace, including aligned blocks,
activation entries, procedure edges, and resource-flow edges.
This makes \sys{} different from a monolithic similarity classifier. It
does not merely say that two skills are similar; it explains where the
provenance evidence appears and which trace survived the reuse
transformation.

\section{Evaluation}
\label{sec:experiments}
\subsection{Research Questions}
We evaluate \sys{} along four research questions.

\begin{itemize}
    \item \textbf{RQ1: Detection effectiveness.}
    Can \sys{} detect skill reuse under realistic rewrites while
    maintaining low false positives on same-function independent skills?

    \item \textbf{RQ2: Trace complementarity.}
    Do Expression, Implementation, and Operational Trace capture
    different reuse patterns, and does the SOG-based Operational Trace
    add evidence beyond text/code similarity?

    \item \textbf{RQ3: Ecosystem discovery.}
    Can \sys{} surface actionable reuse candidates in a real public
    skill corpus, including cases that repository-level baselines
    miss?

    \item \textbf{RQ4: Cost and scalability.}
    Can \sys{} amortize Operational extraction at ingestion
    time and run repeated audits deterministically at marketplace scale?
\end{itemize}


\subsection{Benchmark Construction}
We construct \bench{} from 100 public skill anchors selected from
high-activity skill marketplaces and registries. The anchors cover
diverse task categories, package sizes, and implementation styles,
while limiting duplicates from the same owner and provenance cluster.

\paragraph{Positive reuse pairs}
\bench{} contains 820 positive reuse pairs from these anchors. The
positive set combines observable ecosystem reuse with controlled
transformations. When public provenance is available, we include real
forks, versioned variants, and cross-registry repackages verified by
repository metadata, file paths, names, and package structure. Because
such provenance is sparse and uneven, we also generate controlled
derivatives that model realistic reuse motives. The resulting families
are grouped into the rows reported in Table~\ref{tab:skilltrace-main}.

\begin{itemize}
    \item \textbf{R: Repackaging} models low-effort republication where the
derivative keeps most skill content but changes marketplace-facing
metadata, section layout, or file organization. It contains
R1 metadata rewrite and R2 structural reorganization.
\item \textbf{P: Porting and rewriting} models reuse that preserves the same
skill capability while changing deployment context, host format, or
surface realization. It contains P1 cross-modal port,
P2 registry fork, and P3 full LLM rewrite.
\item \textbf{L: Partial lifting and replacement} models higher-effort reuse
where one valuable part of a skill is preserved while another is
rewritten or replaced. It contains L1
implementation lift, L2 code refactor, and L3
documentation reuse. 
These three regimes span
near-literal reuse, operational migration, and selective reuse of
documentation or implementation.
\end{itemize}

\begin{table*}[t]
\centering
\small
\setlength{\tabcolsep}{4.5pt}
\caption{Per-source coverage on \bench{}. Cells show mean raw similarity with detection rate in parentheses; for each positive row, the best detection rate is bolded. Source columns use $\tau_\lambda$, the 95th percentile on N1 strict negatives. MaxFusion reports the calibrated ratio $\max_i s_i/\tau_i$ and fires at decision bound $\beta=1.0$. The Positive avg. row is a micro-average over all 820 positive pairs. Implementation rates use only applicable pairs where both sides contain sufficient code.}
\label{tab:skilltrace-main}
\begin{tabular}{ll cc cccc}
\toprule
& & \multicolumn{2}{c}{\textbf{Baseline}} & \multicolumn{4}{c}{\textbf{\sys{}}} \\
\cmidrule(lr){3-4} \cmidrule(lr){5-8}
& Subject & Repo-Jac & Repo-ssdeep & Expression & Implementation & Operational & MaxFusion \\
\midrule
\multirow{9}{*}{\rotatebox{90}{\textbf{Positive}}}
& R1 metadata rewrite & \textbf{0.988 (1.00)} & \textbf{0.934 (1.00)} & \textbf{0.982 (1.00)} & \textbf{0.999 (1.00)} & \textbf{0.950 (1.00)} & \textbf{7.174 (1.00)} \\
& R2 structural reorg. & \textbf{0.947 (1.00)} & 0.427 (0.77) & \textbf{0.949 (1.00)} & 0.871 (0.99) & \textbf{0.924 (1.00)} & \textbf{6.578 (1.00)} \\
\cmidrule(l){2-8}
& P1 cross-modal port & 0.203 (0.45) & 0.002 (0.00) & 0.234 (0.62) & \textcolor{gray!55}{0.100 (0.23)} & 0.421 (0.43) & \textbf{1.572 (0.72)} \\
& P2 registry fork & 0.215 (0.48) & 0.000 (0.00) & 0.257 (0.69) & \textcolor{gray!55}{0.084 (0.16)} & 0.498 (0.63) & \textbf{1.797 (0.86)} \\

 & P3 full LLM rewrite & \textcolor{gray!55}{0.156 (0.32)} & 0.009 (0.02) & \textcolor{gray!55}{0.149 (0.31)} & 0.169 (0.24) & 0.681 (0.73) & \textbf{2.724 (0.78)} \\
\cmidrule(l){2-8}
& L1 implementation lift & 0.480 (0.92) & 0.263 (0.47) & \textcolor{gray!55}{0.160 (0.42)} & \textbf{0.835 (1.00)} & \textbf{0.936 (1.00)} & \textbf{6.016 (1.00)} \\
& L2 code refactor & \textbf{0.886 (1.00)} & 0.744 (0.97) & \textbf{0.999 (1.00)} & \textbf{0.806 (1.00)} & \textbf{0.982 (1.00)} & \textbf{6.176 (1.00)} \\
& L3 documentation reuse & 0.389 (0.74) & 0.088 (0.16) & 0.468 (0.88) & 0.196 (0.33) & 0.676 (0.82) & \textbf{3.125 (0.98)} \\
\cmidrule(l){2-8}
& Positive avg. & 0.510 (0.69) & 0.278 (0.37) & 0.517 (0.75) & 0.559 (0.65) & 0.694 (0.76) & \textbf{3.997 (0.88)} \\
\midrule
\multirow{2}{*}{\rotatebox{90}{\textbf{Neg.}}}
& N1 strict negative & \textcolor{gray!55}{0.106 (0.05)} & 0.000 (0.00) & \textcolor{gray!55}{0.092 (0.05)} & \textcolor{gray!55}{0.062 (0.05)} & \textcolor{gray!55}{0.273 (0.05)} & \textcolor{gray!55}{0.786 (0.10)} \\
& N2 clean-room synthesis & \textcolor{gray!55}{0.096 (0.07)} & 0.000 (0.00) & \textcolor{gray!55}{0.103 (0.07)} & \textcolor{gray!55}{0.046 (0.00)} & \textcolor{gray!55}{0.333 (0.10)} & \textcolor{gray!55}{0.856 (0.12)} \\
\cmidrule(l){2-8}
& $\tau_\lambda$ / $\beta$ & 0.182 & 0.000 & 0.172 & 0.139 & 0.406 & 1.000 \\
\bottomrule
\end{tabular}
\end{table*}

\paragraph{Negative controls}
The negative side contains 751 controls and separates reuse from
same-function similarity. N1 contains same-function strict
negatives mined from the real corpus: pairs that solve similar tasks but
are independently implemented. We exclude mirrors, same-owner reuse,
same-provenance clusters, and shared-template-only cases; admission
requires cross-model agreement that the pair is same-function but
independent. N1 is the main false-positive regime for provenance
auditing. N2 contains clean-room skills generated
from task metadata by LLM without access to the original skill body. N2 is not
used for threshold calibration; it is a boundary control for cases where
no body-level trace should be preserved.

\paragraph{Validation}
All benchmark pairs are checked before inclusion. Observable ecosystem
reuse is verified from repository metadata and package structure. For
LLM-generated positives, a cross-model judge whose generator and judge
come from different model families whenever possible is used as an
initial filtering gate. After this filtering step, we manually check the
admitted positive and negative pairs to ensure that the benchmark labels
are reliable. The manual check agrees with the judge on the include/exclude decision for each pair.

\subsection{Experimental Setup}\label{sec:bench}

\subsubsection{Baselines}
We compare against repository-level baselines\cite{kim2025toolcloning},
RepoClone-Jaccard and RepoClone-ssdeep, following the measurement
style of prior tool-cloning studies. RepoClone treats each skill as a
whole package: it flattens repository contents and computes either
token Jaccard similarity or fuzzy-hash similarity with ssdeep~\cite{kornblum2006identifying}. It is
therefore a package-level similarity baseline rather than a
trace-attributed reuse detector. We report the three \sys{} trace
scores separately: Expression, Implementation, and Operational. The
headline method, \sys{}-MaxFusion, surfaces a pair when any calibrated
trace meets the decision bound.


\subsubsection{Evaluation metrics} 
For controlled benchmark evaluation, we report AUROC, precision,
recall, F1, per-family mean scores, and detection rates at thresholds
calibrated only on N1 strict negatives. For the wild-corpus study, which
lacks complete ground truth, we do not report AUROC; instead we report ecosystem
fingdings, disagreement between RepoClone and \sys{}, and manual validation on sampled candidates.

\subsection{RQ1: Detection Effectiveness}

Table~\ref{tab:skilltrace-main} reports per-trace raw similarity and the
thresholded \sys{}-MaxFusion decision score at $\beta=1.0$, broken down by attack family.
The thresholds $\tau_\lambda$ are calibrated only on the N1 strict
same-function negatives. Over the combined negative pool (N1+N2),
\sys{}-MaxFusion reaches AUROC 0.938, precision 0.914, recall
0.883, and F1 0.898, outperforming RepoClone-Max (AUROC
0.841, F1 0.783) and the strongest single trace, Operational
(AUROC 0.878, F1 0.836). Each trace column reports raw similarity:
Jaccard for Expression and Implementation, and aligned SOG-view
coverage for Operational. RepoClone-Max takes the larger calibrated
score of RepoClone-Jaccard and RepoClone-ssdeep.

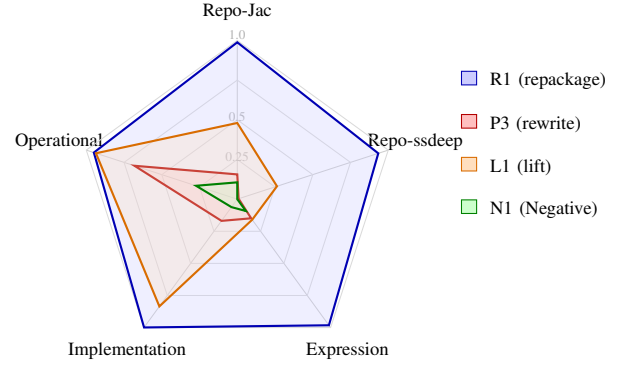
\begin{figure}[t]
\centering
\begin{tikzpicture}[scale=2.1]

\foreach \i/\name in {0/{Repo-Jac}, 1/{Repo-ssdeep}, 2/{Expression}, 3/{Implementation}, 4/{Operational}} {
  \pgfmathsetmacro{\angle}{90 - 72*\i}
  \draw[gray!40, very thin] (0,0) -- (\angle:1.0);
  \node[font=\scriptsize] at (\angle:1.18) {\name};
}

\foreach \r in {0.25, 0.5, 0.75, 1.0} {
  \draw[gray!30, very thin]
    (90:\r) -- (18:\r) -- (-54:\r) -- (-126:\r) -- (162:\r) -- cycle;
}

\node[font=\tiny, gray!60] at (90:0.27) {0.25};
\node[font=\tiny, gray!60] at (90:0.52) {0.5};
\node[font=\tiny, gray!60] at (90:1.04) {1.0};

\draw[blue!70!black, thick, fill=blue!20, fill opacity=0.30]
  (90:0.988) -- (18:0.934) -- (-54:0.982) -- (-126:0.999) -- (162:0.950) -- cycle;

\draw[red!70!black, thick, fill=red!25, fill opacity=0.30]
  (90:0.156) -- (18:0.009) -- (-54:0.149) -- (-126:0.169) -- (162:0.681) -- cycle;

\draw[orange!85!black, thick, fill=orange!25, fill opacity=0.30]
  (90:0.480) -- (18:0.263) -- (-54:0.160) -- (-126:0.835) -- (162:0.936) -- cycle;

\draw[green!50!black, thick, fill=green!20, fill opacity=0.30]
  (90:0.106) -- (18:0.000) -- (-54:0.092) -- (-126:0.062) -- (162:0.273) -- cycle;

\node[font=\scriptsize, anchor=west] at (1.35, 0.75) {
  \tikz \fill[blue!20, draw=blue!70!black] (0,0) rectangle (0.2,0.18); \ R1 (repackage)
};
\node[font=\scriptsize, anchor=west] at (1.35, 0.48) {
  \tikz \fill[red!25, draw=red!70!black] (0,0) rectangle (0.2,0.18); \ P3 (rewrite)
};
\node[font=\scriptsize, anchor=west] at (1.35, 0.21) {
  \tikz \fill[orange!25, draw=orange!85!black] (0,0) rectangle (0.2,0.18); \ L1 (lift)
};
\node[font=\scriptsize, anchor=west] at (1.35, -0.06) {
  \tikz \fill[green!20, draw=green!50!black] (0,0) rectangle (0.2,0.18); \ N1 (Negative)
};

\end{tikzpicture}
\caption{Coverage radar for four representative subjects over the five raw audit metrics in Table~\ref{tab:skilltrace-main}.}
\label{fig:radar_subjects}
\end{figure}

\textbf{Observations.}
Three patterns follow from Table~\ref{tab:skilltrace-main}. First,
R-class repackaging is saturated: near-literal metadata and
structural repackages remain easy for both RepoClone and \sys{}, which
serves as a sanity check for the pipeline. Second, P-class
porting/rewriting is where repository-level
similarity becomes unreliable. In P3 full LLM rewrite, RepoClone-Jaccard
drops to 0.156 with only 0.32 detection rate, while Operational
remains the strongest single trace at 0.681 mean similarity and
0.73 detection rate; MaxFusion raises the P3 detection rate to
0.78. This reflects the key skill-specific signal: even when wording
and code are rewritten, a functional rewrite tends to preserve the
operational procedure needed for the skill to work. Third,
L-class partial lifting/replacement transformations show
that the firing trace rotates with the reuser's preservation choice.
In L1 implementation lift, Expression collapses to 0.160, but
Implementation rises to 0.835 and Operational to 0.936; in L3
documentation reuse, Expression becomes the dominant surface trace.
Across L1--L3, MaxFusion fires on at least 0.96 of positives.

Figure~\ref{fig:radar_subjects} visualizes the same pattern. R1 fills the
outer ring because all traces are preserved. P3 collapses on repository
and lexical axes but retains the Operational trace. L1 suppresses
Expression while preserving Implementation and Operational trace. N1
strict negatives remain near the center, showing that the calibrated
thresholds are not simply detecting same-function similarity.

\begin{tcolorbox}[fonttitle=\bfseries, boxsep=3pt, left=6pt, right=6pt, top=2pt, bottom=2pt]
\textbf{Answer to RQ1:} \sys{} improves over repository-level
baselines and single-trace variants on realistic skill reuse, reaching
AUROC 0.938 and F1 0.898.
\end{tcolorbox}

\subsection{RQ2: Trace Complementarity}

A natural concern is whether the three traces in \sys{} are merely
different names for the same surface similarity. We do not claim that
the traces are statistically independent features: near-literal
repackages should make all traces rise together. Instead, RQ2 asks
whether the traces provide complementary forensic paths when a reuse
transformation rewrites one part of a skill while preserving another. We
examine this at three levels: score correlation, thresholded firing
patterns, and fusion behavior.

\textbf{Trace-level complementarity.}
Table~\ref{tab:source_corr} reports Pearson correlations between
Expression, Implementation, and Operational similarities. The useful
signal is not the all-positive average, which mixes easy repackages
with harder rewrites, but the family-wise split. In partial-lift
reuses, Expression--Implementation correlation falls to $r{=}0.15$:
some derivatives preserve executable material while rewriting
documentation, while others preserve documentation and replace code. In
porting and rewriting reuses, Expression--Operational and
Implementation--Operational correlations drop to $r{=}0.28$ and
$r{=}0.36$. This is the regime where Operational Trace matters most:
words and code can be rewritten, but a useful derivative often preserves
the procedure, tool roles, and execution structure needed for the skill
to work.

\begin{table}[t]
\centering
\small
\caption{Pearson correlation between trace similarities by reuse family.}
\setlength{\tabcolsep}{8pt}
\begin{tabular}{@{}lccc@{}}
\toprule
Reuse Family & E--I & E--O & I--O \\
\midrule
All positives       & 0.76 & 0.62 & 0.67 \\
R repackage         & 0.41 & 0.13 & 0.43 \\
P port / rewrite    & 0.43 & 0.28 & 0.36 \\
L partial lift   & 0.15 & 0.20 & 0.72 \\
\bottomrule
\end{tabular}

\label{tab:source_corr}
\end{table}

\textbf{Decision-level complementarity.}
Thresholded decisions show the same effect. Under the default
$\beta=1.0$ bound, a non-trivial set of positives fires through exactly
one trace: about one hundred Expression-only cases, about seventy
Operational-only cases, and only a few Implementation-only cases. The
Operational-only cases are especially important: they are mostly
cross-modal ports and full LLM rewrites where repository overlap,
authored wording, and code tokens fall below threshold, but the
operational procedure remains detectable. Implementation-only fires are
rare because reused implementation usually also carries the procedure
that invokes it. The point is therefore complementarity rather than
orthogonality: each trace provides a different path by which preserved
provenance can surface.

\begin{table}[t]
\centering
\caption{Global detection effectiveness against the negative pool. Scores are trace-normalized by N1-calibrated thresholds and evaluated at decision bound $\beta=1.0$; linear fusion averages applicable normalized traces. Implementation-only is evaluated on its applicable subset.}
\small
\resizebox{\columnwidth}{!}{%
\begin{tabular}{@{}lrrrr@{}}
\toprule
Variant & AUROC & Precision & Recall & F1 \\
\midrule
RepoClone-Max (baseline) & 0.841 & 0.924 & 0.679 & 0.783 \\
Expression only & 0.866 & 0.939 & 0.750 & 0.834 \\
Implementation only & 0.744 & 0.935 & 0.647 & 0.765 \\
Operational only & 0.878 & 0.934 & 0.756 & 0.836 \\
Linear fusion (equal E/I/O) & 0.909 & \textbf{0.973} & 0.744 & 0.843 \\
\textbf{\sys{}-MaxFusion} & \textbf{0.938} & 0.914 & \textbf{0.883} & \textbf{0.898} \\
\bottomrule
\end{tabular}}

\label{tab:global-detection}
\end{table}

\textbf{Decision-rule ablation.}
Table~\ref{tab:global-detection} compares single-trace variants, linear
fusion, and \sys{}-MaxFusion against the combined N1+N2 negative pool.
Linear fusion over normalized E/I/O traces is conservative: it reaches
the highest precision (0.973) but lower recall (0.744), because a
strong trace is averaged against weaker or rewritten traces. MaxFusion
instead surfaces a pair when any calibrated trace fires, giving the best
AUROC, recall, and F1 while keeping precision high (0.914). This
supports the forensic design choice behind MaxFusion: localized
provenance evidence should be sufficient to route a pair for review,
while the report still identifies which trace fired.

\begin{tcolorbox}[fonttitle=\bfseries, boxsep=3pt, left=6pt, right=6pt, top=2pt, bottom=2pt]
\textbf{Answer to RQ2:} The three traces are complementary forensic
paths rather than redundant aliases. Expression, Implementation, and
Operational traces split under different reuse scenarios, and
MaxFusion preserves localized evidence that linear averaging would
dilute.
\end{tcolorbox}

\subsection{RQ3: Ecosystem Discovery}\label{sec:realworld}

We audit a public corpus of 36,446 real-world skills collected from
SkillHub and ClawHub as a deployment study. The corpus has no complete
provenance ground truth, so RQ3 is not a supervised benchmark: we do
not report AUROC on wild pairs and we do not make legal infringement
claims. Instead, we ask whether \sys{} can build actionable,
trace-attributed reuse review queues at marketplace scale. We run two
complementary audits: an anchor-driven scan over popular skills and a
global audit over the full corpus.


\vspace{1mm}
\textbf{Top-50 hot-skill exposure.}
We first audit from the perspective of popular skill authors by taking
the top 50 skills from SkillHub as anchors and
searching the full corpus for conservative reuse candidates. The result
shows that reuse around popular skills is widespread but uneven: 49 of
the 50 hot anchors have at least one jointly flagged reuse candidate
where both RepoClone and \sys{} evidence fire, and the scan
surfaces 357 such candidates in total. Reuse is concentrated rather
than uniformly distributed: 7 anchors have more than ten reuse
candidates, with \texttt{self-improvement} alone surfacing 41.
Software-engineering skills account for the largest volume, with 211
reuse candidates across 30 anchors.

The anchor scan also exposes the regime that motivates trace attribution.
Static \sys{} surfaces 18 \sys{}-only candidates across 8 anchors:
repository-level similarity stays below the conservative threshold, but
\sys{} evidence is strong enough to justify human
review. Thus, broad full-skill reuse is the dominant hot-anchor pattern,
while source-specific partial reuse is rarer but visible and would be
easy to under-prioritize with package-level similarity alone.

\vspace{1mm}
\textbf{Global reuse audit.}
We next scale from hot anchors to the full corpus. A naive all-pair
audit would require more than $6.6\times10^8$ comparisons, so \sys{}
uses source-indexed candidate generation: it builds indexes over
repository tokens, Expression tokens, and Implementation tokens,
enumerates pairs that share enough static evidence to plausibly pass a
conservative review threshold, unions candidates across sources, and
then exact-scores the resulting root-pair set. RepoClone serves as the
package-level baseline, while \sys{} records which static trace fired.
We deliberately keep Operational Trace out of the global routing rule:
it is useful as corroborating evidence, but generic same-function skills
can collide on common operational schemas such as API wrappers or
document-conversion pipelines.

The global pass exact-scores $204{,}386$ deduplicated candidate pairs.
On this pool, RepoClone and static \sys{} evidence jointly flag
$43{,}581$ pairs ($21.3\%$), the broad-reuse region where package-level
and trace-level evidence agree. \sys{} additionally surfaces $3{,}030$
static-only candidates ($1.5\%$): $1{,}529$ Expression-only, $1{,}338$
Implementation-only, and $163$ where both static traces fire while
RepoClone remains below threshold. These pairs are the main ecosystem
contribution of \sys{}: they preserve inspectable source-level evidence
without enough whole-package overlap to dominate a repository-level
queue. The reverse region is much smaller: $439$ pairs ($0.2\%$) are
RepoClone-only, a mixed bucket of scaffolds, templates, and borderline
reuse. Overall, near-full reuse is more common in the wild, but partial
source-level reuse appears at nontrivial scale.



\vspace{1mm}
\textbf{Manual validation.}
The automatic buckets are triage queues, not provenance verdicts: they
decide what a marketplace auditor should inspect, not whether
infringement occurred. We therefore blindly reviewed 200 stratified
wild pairs, including 100 pairs surfaced by \sys{} and 100
low-priority controls. The \sys{}-flagged pairs consistently contain
inspectable reuse evidence aligned with the firing trace: reused
documentation, examples, support scripts, or
implementation fragments. The low-priority controls are overwhelmingly
independent or unclear and lack comparable provenance evidence. This
manual check supports the central wild-corpus claim: \sys{}'s
source-attributed queues correspond to evidence a human reviewer can
inspect, while the tool itself stops short of legal adjudication.

\begin{tcolorbox}[fonttitle=\bfseries, boxsep=3pt, left=6pt, right=6pt, top=2pt, bottom=2pt]
\textbf{Answer to RQ3:} In a large public corpus, \sys{}
builds actionable reuse review queues beyond package-level similarity:
hot-skill reuse is widespread, and global auditing
surfaces a substantial set of trace-attributed candidates that
repository-level baselines would under-prioritize.
\end{tcolorbox}

\subsection{RQ4: Cost and Scalability}
\label{sec:cost}

\sys{} separates per-skill registration from repeated pairwise audits.
At ingestion time, Expression and Implementation traces are extracted
deterministically, while the Operational trace is extracted once with an
LLM and cached with its prompt version. In our configuration, cold
Operational extraction costs about 26 seconds and \$0.01 per skill
(roughly 10K input and 1K output tokens). Registering the
36,446-skill wild corpus therefore costs about \$360 and 263
single-worker hours, or 5.3 hours with 50 parallel workers.

After registration, audit-time scoring uses only cached traces and
makes no LLM calls. On a cached pairwise timing sample, full
\sys{}-MaxFusion scoring takes 105 ms per pair on average. Even using
this full-score timing as a conservative bound, the 204,386 global
candidate pairs in RQ3 require about 6.0 single-worker hours, or
roughly 22 minutes with 16 workers, and zero token cost. Thus
\sys{}'s cost is paid once when a skill enters the registry; repeated
reuse audits are deterministic and inexpensive.

\begin{tcolorbox}[fonttitle=\bfseries, boxsep=3pt, left=6pt, right=6pt, top=2pt, bottom=2pt]
\textbf{Answer to RQ4:} \sys{} amortizes LLM use at ingestion time.
After traces are cached, marketplace-scale audits run deterministically
with zero token cost, and minutes-scale latency
under modest parallelism.
\end{tcolorbox}



\subsection{Threats to validity.} 

The external threats to validity mainly come from benchmark construction and corpus
coverage. Some positive pairs are generated by LLM-assisted
transformations, which may introduce model-specific style. We reduce
this risk through cross-model generation and judging, per-trace
reporting, and manual spot checks, but the benchmark cannot cover every
future marketplace reuse pattern. The wild audit also covers public
registries rather than all commercial or private skill ecosystems; we
therefore report it as a deployment study and review-queue analysis, not
as a prevalence estimate for all skill markets.

Internal threats concern trace extraction and scoring correctness.
\sys{} uses LLM assistance only for Operational trace extraction at
ingestion time; audit-time scoring is deterministic over cached traces.
The extractor runs under an auditor-controlled, temperature-zero,
versioned prompt, but extracted SOGs can still be noisy or too coarse.
This matters especially for generic same-function skills. For this reason, Operational evidence is calibrated against strict negatives and used cautiously in wild-corpus routing. Similarly,
Implementation Trace is applicable only when both skills contain enough
code or command material; code-light skills should not be treated as
implementation mismatches.

Construct threats concern what a trace match means. \sys{} measures
provenance evidence, not legal infringement, intent, license
compatibility, or ownership. Shared templates, common API wrappers, and
multi-origin merges can produce genuine similarity without a clean
one-to-one reuse relation. We therefore surface trace-attributed review
candidates rather than automatic verdicts. Reuse that deliberately
targets the detector is likewise outside our current benchmark; as
skill ecosystems evolve, benchmark families, strict negatives, and
human-review protocols will need to evolve with them.

\section{Discussion}\label{sec:discussion}
\subsection{Registry deployment and governance}
\sys{} is designed for marketplace-scale deployment: skills pay a
one-time registration cost for trace extraction, while repeated audits
operate over cached substrates with no audit-time LLM call
(\S\ref{sec:cost}). This separation lets registries reuse the same
traces for dispute review, duplicate management, and quality ranking.
A registry can use source-attributed reuse evidence to cluster related
skills, surface a more complete or maintained variant, link
alternatives, and warn users about smaller or outdated derivatives. The
trace labels make this more actionable than a monolithic score:
Expression points to reused prose or examples, Implementation to lifted
scripts or command/API patterns, and Operational to preserved
agent-facing behavior.

\subsection{Protectability and legal boundary}
We use protectability only as a review-prioritization concept. Skills
with substantial authored instructions, reusable implementation,
distinctive tool/resource organization, or nontrivial operational design
are more worth auditing because these properties reflect author effort
and marketplace value. \sys{} does not decide legal protectability; it
identifies where provenance evidence survives and what deserves human
review. More broadly, \sys{} produces provenance evidence, not legal
verdicts. It does not infer intent, ownership, license compatibility, or
infringement; those depend on jurisdiction, license terms, fair use, and
marketplace policy. Accordingly, wild-corpus reports use anonymized
cases and describe surfaced pairs as review candidates. The tool's role
is forensic triage: which traces survived, where to inspect, and whether
follow-up is warranted.

\subsection{Operational trace and skill value}
Operational Trace is the most skill-specific trace in \sys{}: it acts
as a skill-level operational birthmark. Yet Operational similarity
alone is not a reuse verdict. Same-function skills may collide on
generic operational schemas because functional requirements can impose
similar high-level procedures. This is an information-content issue
rather than a failure of the trace: for complex skills, a richer SOG
carries more distinctive design decisions and can serve as a stronger
birthmark; generic, low-information SOG patterns should not be treated
as reuse evidence by themselves. We therefore treat generic schema
matches as an ambiguous review zone rather than a positive finding.
Operational evidence is strongest when corroborated by Expression,
Implementation, distinctive resources, or propagation patterns. Beyond
provenance auditing, the SOG may also provide a future signal for
skill-value assessment, since richer operational structure can indicate
more substantial agent-facing design.
\vspace{1mm}

\section{Conclusion}\label{sec:conclusion}

In this work, we propose \sys{}, a multi-trace provenance auditing framework for agent skills. To capture provenance signals under realistic skill reuse, \sys{} adopts a two-stage design. The trace extraction stage extracts Expression, Implementation, and Operational traces, with the Operational Trace instantiated as a Skill Operational Graph. The reuse detection stage compares these traces separately with calibrated thresholds, allowing \sys{} to surface not only whether a candidate is suspicious but also which trace supports the decision. Experiments on \bench{} and a large public skill corpus demonstrate the effectiveness, interpretability, and deployment practicality of \sys{} in reuse auditing. Compared with baselines, \sys{} achieves stronger detection performance while surfacing trace-attributed review evidence that global package similarity misses. A promising direction for future work is to reuse these trace representations for adjacent marketplace tasks, such as skill quality assessment and trace-aware skill search.

\bibliographystyle{IEEEtran}
\bibliography{references}


\end{document}